\documentclass{article}
\usepackage{ijcai24}

\usepackage{times}
\usepackage{soul}
\usepackage{url}
\usepackage[hidelinks]{hyperref}
\usepackage[utf8]{inputenc}
\usepackage[small]{caption}
\usepackage{graphicx}
\usepackage{amsmath}
\usepackage{amsthm}
\usepackage{booktabs}
\usepackage{algorithm}
\usepackage{algorithmic}
\usepackage[switch]{lineno}
\usepackage{multirow}
\usepackage{amsfonts,amssymb}

\title{Visual Watermarking in the Era of Diffusion Models: Advances and Challenges}

\author{
Junxian Duan$^1$
\and
Jiyang Guan$^1$\and
Wenkui Yang$^1$\And
Ran He$^1$\\
\affiliations
$^1$State Key Laboratory of Multimodal Artificial Intelligence Systems, CASIA, Beijing, China\\
\emails
\{junxian.duan, guanjiyang2020\}@ia.ac.cn,
wenkui.yang@cripac.ia.ac.cn,
rhe@nlpr.ia.ac.cn
}

\begin{document}

\maketitle

\begin{abstract}
As generative artificial intelligence technologies like Stable Diffusion advance, visual content becomes more vulnerable to misuse, raising concerns about copyright infringement. Visual watermarks serve as effective protection mechanisms, asserting ownership and deterring unauthorized use. Traditional deepfake detection methods often rely on passive techniques that struggle with sophisticated manipulations. In contrast, diffusion models enhance detection accuracy by allowing for the effective learning of features, enabling the embedding of imperceptible and robust watermarks. We analyze the strengths and challenges of watermark techniques related to diffusion models, focusing on their robustness and application in watermark generation. 
By exploring the integration of advanced diffusion models and watermarking security, we aim to advance the discourse on preserving watermark robustness against evolving forgery threats. It emphasizes the critical importance of developing innovative solutions to protect digital content and ensure the preservation of ownership rights in the era of generative AI.

\end{abstract}

\section{Introduction}


In the burgeoning age of generative AI, watermarks
act as identifiers of provenance and artificial content \cite{An2024bench}. Visual watermarking involves embedding recognizable patterns or logos in digital media to assert ownership and prevent unauthorized use. The main deepfake detection methods typically identify deepfake images in a passive manner by extracting biometric features and high-frequency artifacts from the images \cite{nature24}. However, as deepfake images become more realistic or are transmitted through lossy channels, passive detection may fail to identify them, limiting their effectiveness. As the demand for digital content increases, so does the need for effective protection mechanisms. Visual watermarks deter infringement and help trace unauthorized reproductions, making them essential for digital content security \cite{Bao0WLWJC24}.

Diffusion models (DMs) \cite{sohl2015deep,ho2020denoising,rombach2022high} have gained significant traction in the fields of image generation and recognition, offering innovative solutions for both creative and security challenges. Diffusion models transform noise into coherent visual content through a learned denoising process, which has made them a leading tool for applications such as artistic creation and data augmentation \cite{Liyifan23}. In forgery detection, diffusion models leverage their deep understanding of data distributions to identify inconsistencies and anomalies in images \cite{diffforgery24}. By combining their generative capabilities with robust analytical features, diffusion models provide a powerful framework for both creating and safeguarding digital media.


The increasing popularity of open-source Stable Diffusion models has led to serious concerns about intellectual property and verification, impacting both data contributors and the creators of these models. To address these issues, Watermarking techniques have emerged as a promising method to ensure traceability and protect intellectual property \cite{zhao2024SoK}. Current methods include data-driven passive approaches (embedding watermarks into training data or fine-tuning pre-trained models) \cite{zhao2023recipe}, direct modification of DMs’ sampling strategies or parameters (altering output distributions for traceability) \cite{wen2023tree}, and adversarial watermarking (embedding invisible adversarial marks to prevent unauthorized data usage) \cite{liang2023adversarial}. These solutions aim to mitigate misuse while balancing the challenges and opportunities introduced by the widespread use of DMs. 
Meanwhile, the integration of watermarking and acceleration is non-trivial \cite{HuH24}, requiring careful consideration of trade-offs between efficiency and robustness. Despite these advancements, current reviews tend to analyze visual watermarking and diffusion models in isolation, either focusing on the contributions of diffusion models to content protection or examining watermarking techniques within diffusion models, which overlooks the critical interplay between the two domains. Given that diffusion models may introduce unique challenges related to robustness and security, there is an urgent need for a comprehensive examination of their relationship with visual watermarking techniques.


This paper seeks to examine the applications and potential of diffusion model-based visual watermarking for enhancing the security of generative content. By focusing on this innovative approach, we aim to demonstrate how diffusion models can improve the effectiveness of watermarking in an increasingly complex digital environment. Besides, we will analyze the strengths and challenges associated with utilizing diffusion models for watermark generation, embedding, and robustness. Through this examination, we intend to provide a thorough understanding of how these models can advance watermarking techniques while addressing the critical issues of maintaining watermark integrity against various forms of digital manipulation. This study distinguishes itself by exploring the convergence of cutting-edge generative methods and watermarking security, a topic that has received limited attention in the current literature.


To begin, Section \ref{pre} explores the preliminaries of visual watermarking and diffusion models, highlighting their unique characteristics and advantages. Section \ref{diffusion} introduces the watermarking in diffusion models by passive watermarking and proactive watermarking, including effectiveness and robustness. In Section \ref{challenge}, we address challenges like computational costs, attack against, and the need for multi-attribution. Section \ref{application} showcases applications in copyright protection, forensics, and data privacy. Finally, Section \ref{conclusion} concludes with future directions and implications. In summary, although watermarking may not fully resolve the issues of visual content misuse, it serves as an essential mechanism for improving multimedia information security. This paper aims to offer a perspective on the dynamic evolution of watermarking techniques in response to emerging generative models, emphasizing important opportunities and challenges for the trustworthy and secure implementation of visual watermarking in visual content protection.

\section{Preliminary} \label{pre}
To provide essential context for understanding the interplay and challenges, we first introduce visual watermarking, covering its methods and applications for content protection. Then, we explore diffusion models, emphasizing their unique architecture or training mechanisms, and highlighting their architecture and role in generative AI. By examining these two areas in tandem, this section establishes a comprehensive foundation for analyzing their integration.

\subsection{Visual Watermarking Overview}
\textbf{Visual watermarking} \cite{cox1997secure,petitcolas1999information} embeds information directly into visual content to safeguard copyrights, authenticate content, and verify data integrity. In practice, watermarking techniques generally fall into two categories: robust and fragile. Robust watermarks are engineered to withstand common signal processing operations (e.g., compression, cropping, or the addition of noise) and malicious manipulation, ensuring that the watermark remains detectable even after such alterations. In contrast, fragile watermarks are designed to be extremely sensitive to any changes; even the slightest modification can disrupt them, making these watermarks perfect for verifying that content has not been tampered with.

Methodologically, visual watermarks can be embedded either in the spatial domain, where pixel or sample values are directly modified, or in the frequency domain, where information is inserted into the transformed coefficients using techniques like the Discrete Cosine Transform (DCT) or Wavelet Transform (DWT). While spatial domain methods like Least Significant Bit (LSB) are straightforward, they generally offer less resistance to tampering compared to frequency domain approaches, which provide greater robustness against various signal processing operations. 

More recently, the emergence of deep neural networks has revolutionized watermarking by enabling end-to-end learning frameworks that automatically optimize both the embedding and extraction processes \cite{zhu2018hidden,zhang2019robust,luo2020distortion,zhang2020udh}, and it is also inherently compatible with blind watermarking approaches, where the extraction of the watermark does not require access to the original content. For example, HiDDeN \cite{zhu2018hidden} firstly incorporates encoder-decoder networks to embed and extract watermarks with high resilience to distortions such as compression, noise, and geometric transformations.

However, with the advancement of generative models, such as GANs and DMs, as well as text-to-image (T2I) models, watermarking techniques are increasingly challenged by issues of insufficient robustness. On the one hand, watermarked contents may encounter more complicated workflows due to the popular large-scale pretraining scheme, rendering robustness an increasingly challenging issue. On the other hand, the high computational and financial costs associated with pretrained models make them also prime targets for malicious appropriation. This necessitates the introduction of novel generative techniques.

\begin{figure}[h]
\centering
\includegraphics[width=0.97\linewidth]{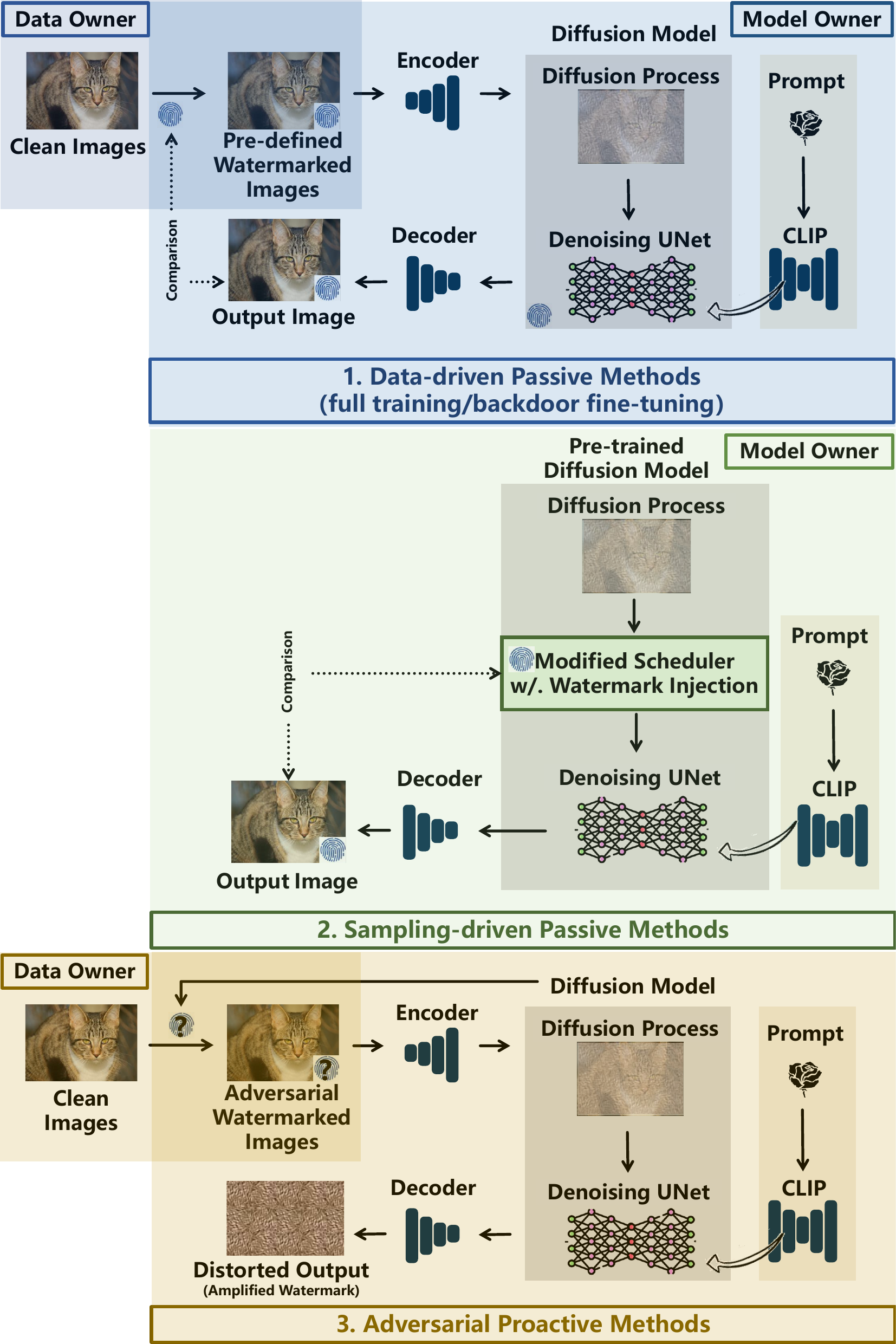}
\caption{
Three common workflows for DMs watermarking.
(Top) Data-driven passive methods require specific training data to be pre-embedded with watermarks, allowing the watermark to be transferred from data into the DMs. 
(Middle) Sampling-driven passive methods modify DMs' sampling strategies, altering the output distribution to embed watermarks.
(Bottom) Adversarial proactive methods apply adversarial attacks against DMs to introduce adversarial watermarks, leading to unrecognizable examples that hinder effective customized outputs.
}
\label{fig:workflow}
\end{figure}

\subsection{Diffusion Models: A New Paradigm}

Since their introduction, Diffusion Models \cite{sohl2015deep,ho2020denoising} have emerged as a paradigm in generative modeling, owing to their exceptional distribution fitting capabilities. The core principle of DMs lies in mapping between an unknown, complex data distribution and a Gaussian distribution through approximately inverse diffusion and denoising processes. Specifically, the forward diffusion process is implemented using a Markov diffusion kernel: 
\begin{equation}
q(\mathbf{x}_t \ | \ \mathbf{x}_{t-1}) = \mathcal{N}(\mathbf{x}_t; \sqrt{1 - \beta_t} \mathbf{x}_{t-1}, \beta_t \mathbf{I}), \  t = 1, 2, \ldots, T
\label{eq:DMforward1}
\end{equation}
where \( t \) represents the timesteps, and \( \mathbf{x}_0 \sim q(\mathbf{x}_0) \) denotes the target distribution to be learned. After \( T \) diffusion steps, the marginal distribution reaches \( q(\mathbf{x}_T) \approx \mathcal{N}(\mathbf{x}_T; \mathbf{0}, \mathbf{I}) \). The coefficient \( \beta_t \in (0,1) \) controls the variance at each step. This diffusion kernel further allows for a closed-form expression of the conditional distribution at any step \( t \), given \( \mathbf{x}_0 \):
\begin{equation}
q(\mathbf{x}_t \ | \ \mathbf{x}_{0}) = \mathcal{N}(\mathbf{x}_t; \sqrt{\overline{\alpha}_t} \mathbf{x}_{0}, (1 - \overline{\alpha}_t) \mathbf{I})
\label{eq:DMforward2}
\end{equation}
where $\quad \overline{\alpha}_t := \prod_{i=0}^t (1 - \beta_i)$. Thus, with Bayes' theorem and the Markov property of the diffusion kernel, we have:
\begin{equation}
q(\mathbf{x}_{t-1} \ | \ \mathbf{x}_{t}, \mathbf{x}_{0}) = \frac{q(\mathbf{x}_{t} \ | \ \mathbf{x}_{t-1}) \cdot q(\mathbf{x}_{t-1} \ | \ \mathbf{x}_{0})}{q(\mathbf{x}_{t} \ | \ \mathbf{x}_{0})}
\label{eq:DMforward3}
\end{equation}
Thus, DDPMs' \cite{ho2020denoising} training objective is to align the model distribution $p_\theta(\mathbf{x}_{t-1}|\mathbf{x}_{t})$ with the known forward distribution $q(\mathbf{x}_{t-1} | \mathbf{x}_{t}, \mathbf{x}_{0})$, thereby learning an optimal denoising strategy. This is achieved by minimizing the difference between the means of two Gaussian distributions, which can be further simplified as:
\begin{equation}
\mathcal{L}_{ddpm} = \mathbb{E}_{\mathbf{x}_0, \epsilon \sim \mathcal{N}(0,1), t \sim \mathcal{U}(1,T)}
\left\| \epsilon - \epsilon_{\theta}(\mathbf{x}_t(\mathbf{x}_0, t), t) \right\|_2^2
\label{eq:DDPM}
\end{equation}

Latent Diffusion Models (LDMs) \cite{rombach2022high} employ an encoder \( \mathcal{E}(\cdot) \) to map the input image into a latent representation \( \mathbf{z}_0 = \mathcal{E}(\mathbf{x}_0) \). The above diffusion-denoising processes are conducted within this low-dimensional latent space, with the introduction of an encoded prompt condition $\tau_\theta(y)$. The training objective is then formulated as:
\begin{equation}
\mathcal{L}_{ldm} = \mathbb{E}_{\mathbf{z}_0, y, \epsilon \sim \mathcal{N}(0,1), t \sim \mathcal{U}(1,T)} 
\left\| \epsilon - \epsilon_{\theta}(\mathbf{z}_t, t, \tau_{\theta}(y)) \right\|_2^2
\label{eq:LDM}
\end{equation}

The widely adopted open-source Stable Diffusion (SD) models are implemented based on LDM architecture. However, this high degree of customization introduces significant challenges related to copyright and authentication. On the one hand, image data providers are concerned that unseen models may learn from and misuse their work without proper authorization. Even in the absence of ethical concerns regarding data usage, model developers similarly seek to prevent their models from being commercially exploited or employed for malicious purposes without consent.

These issues can be mitigated through DMs watermarking to identify model and data provenance. In this context, DMs are not solely responsible for generating desired visual content but also for embedding watermarks for traceability purposes. As illustrated in Figure \ref{fig:workflow}, current DMs watermarking methods can be categorized into three main approaches:

The first category is data-driven passive methods, which build on traditional watermarking techniques and indirectly influence DMs' outputs by embedding watermarks into the training data. More specifically, these methods can be further subdivided into full training methods \cite{zhao2023recipe} and backdoor attack fine-tuning methods \cite{liu2023watermarking}. Full training methods are suitable for organizations capable of supporting large-scale text-to-image model training. In this approach, model-specific watermarks are embedded into the entire pretraining dataset, ensuring that all outputs can be traced back for model IP provenance. Similarly, data owners can also embed watermarks into their content, enabling traceability for specific subsets of training data.
Backdoor attack fine-tuning methods apply to models that have already been pre-trained on unwatermarked clean data. These methods involve fine-tuning with predefined watermarked image-prompt pairs, allowing the model's origin to be verified by determining whether a trigger prompt successfully elicits the expected watermarked output.

Another common approach involves direct modification of DMs’ sampling strategies or internal parameters \cite{ci2024ringid,kim2024wouaf}. While also applying to pre-trained models, these methods differ from data-driven ones by directly inducing shifts in the model’s output distribution. Thus, a given output can be traced by verifying whether it comes from a model employing a specific watermarking sampling strategy. The sampling protocol used for traceability can be defined by a predefined authentication key.

The final category targets data owners rather than model owners. These users aim to prevent their published data from being exploited for training or customized outputs without authorization. To achieve this, adversarial attacks can be applied against potentially unseen DMs to embed invisible adversarial watermarks into the data \cite{liang2023adversarial,van2023anti}. This process renders the data as unrecognizable examples, thereby obstructing effective model outputs and protecting the data from unauthorized usage.

Overall, the widespread adoption of DMs has introduced new challenges for watermarking, while simultaneously offering a broader range of potential solutions.

\begin{table*}\footnotesize 
    \centering
    \begin{tabular}{lll}
        \hline
        Category  & Method & Description \\
        \hline
        \multirow{6}{*}{PW for Image}  &  HiDDeN \cite{zhu2018hidden} & It directly adds watermarks to generated content. \\
        & Stable Signature \cite{fernandez2023stable}     & It fine-tunes the decoder of LDMs to embed a binary signature. \\
        & Tree-Ring \cite{wen2023tree} & It introduces watermarks in the Fourier space. \\
        & RingID \cite{ci2024ringid} & It strengthens multi-key recognition with multi-channel watermarking.\\
        & WOUAF \cite{kim2024wouaf} & It watermarks by modulating the parameters of each layer in the decoder.\\
        & GrIDPure \cite{zhao2024can} & It discloses the vulnerability of existing watermarking methods.\\
        \hline
        \multirow{2}{*}{PW for Model} & FixedWM/NaiveWM \cite{liu2023watermarking} & It backdoors models with a designated keyword and a target image. \\
        & Recipe \cite{zhao2023recipe} & It fine-tunes pre-trained diffusion using trigger pairs.\\
        \hline
        \multirow{5}{*}{PAW for Image} & 
        advDM \cite{liang2023adversarial} &   It utilizes PGD to maximize the training loss of the SD. \\
         & Glaze \cite{shan2023glaze} & It protects artistic IP by perturbing the encoding process of SD.\\
         & Anti-DB  \cite{van2023anti} &  It optimizes noise on surrogate models to defend against DreamBooth.\\
         & MetaCloak \cite{liu2024metacloak} & It enhances Anti-DreamBooth by integrating meta-learning techniques.  \\
         & SIMAC \cite{wang2024simac} & It enhances with regularization on latent space and small time steps.\\ 
         \hline

    \end{tabular}

    \caption{Representive Paper List for Diffusion Watermarking. PW represents passive watermarks and PAW represents proactive watermarks.}
    \label{tab:plain}
\end{table*}

\section{Diffusion Models for Visual Watermarking} \label{diffusion}

With the advancement of diffusion models, they have been applied in many critical fields. However, the powerful generative and editing capabilities of diffusion models have also raised concerns about their potential misuse, as well as the abuse of other AIGC-based generative models. To address these concerns, watermarking has been proposed as a solution. By adding small perturbations to images while minimizing the impact on image quality, watermarking helps protect the security, privacy, and intellectual property of both models and data.
We summarize the representive works of watermarks in diffusion models in three categories in Table \ref{tab:plain}. To be specific, PW for Image and PW for Model represent passive watermarks for detecting the IP of images(generated content) and well-trained models, respectively.

\subsection{Watermarking in Diffusion Models}

Overall, model watermarking can be categorized into proactive watermarking and passive watermarking based on its purpose. Passive watermarking involves adding small noise to images to enable copyright tracking, detection, and provenance verification of images or models. In contrast, proactive watermarking utilizes adversarial optimization to introduce noise into images, preventing third parties from modifying or editing the data using diffusion models.

\paragraph{Passive Watermarking.}

Passive watermarking has been proposed for many years and has been used to detect multimedia content. Invisible watermarks are added to images or text, allowing these contents to be detected.
With the development of diffusion models, the richness of generated content has drawn much attention. Both the copyright of the generated content and the copyright of the models require the introduction of watermarking-related methods into the diffusion area.
At the same time, considering the computational overhead requirements of diffusion model watermarking and the demand for model open-sourcing, mainstream diffusion model watermarking methods aim to integrate the watermarking mechanism into the inference process of diffusion models.
A typical example Stable Signature \cite{fernandez2023stable}, which combines traditional watermarking methods, employs the classic HiDDeN \cite{zhu2018hidden}, and fine-tunes the decoder of LDMs to embed a binary signature.
Specifically, the loss of finetuning the decoder networks as follows:
\begin{equation}
    L = L_{m} + \lambda L_{i} = BCE(\sigma(m^{'},m)) + L_{i}
\end{equation}
where $L_{m}$ represents the message loss to detect the signatures with a BCE loss and $L_{i}$ represents the image perception loss.
With the training loss for the decoder, the watermark is embedded into the decoder parameters of the diffusion model.
Furthermore, compared to Stable Signature \cite{fernandez2023stable}, which primarily embeds the watermark into the diffusion model's decoder, Tree-Ring \cite{wen2023tree} takes a deeper step into the denoising phase by introducing ring patterns in the Fourier space of the initial noise vector. This enables the embedding of watermark information at the distributional level rather than merely at the sample level.

The above primarily focuses on the application of passive watermarking in detecting generated images. Since training models require collecting large amounts of data and consuming significant computational resources, models themselves are considered valuable assets. As a result, model copyright protection has also gained widespread attention.
To solve the problem mentioned above, NaiveWM and FixedWM \cite{liu2023watermarking} have been proposed to backdoor models with a designated keyword and a specific target image.
Recipe \cite{zhao2023recipe} and WDP \cite{peng2025intellectual} also propose to fine-tune pre-trained diffusion using trigger pairs.

\paragraph{Proactive Watermarking.}
Malicious image tampering has long been a critical research problem. The proliferation of image editing methods such as DreamBooth has raised concerns about unauthorized modifications. To address this, methods like AdvDM \cite{liang2023adversarial}, Mist \cite{liang2023mist}, Glaze \cite{shan2023glaze}, and PhotoGuard \cite{salman2023raising} leverage adversarial example generation techniques, such as Projected Gradient Descent (PGD) \cite{madry2017towards}, to maximize the training loss of the Latent Diffusion Model, thereby preventing unauthorized edits.
SIMAC \cite{wang2024simac} and MetaCloak \cite{liu2024metacloak} enhance Anti-DreamBooth by integrating meta-learning techniques and introducing additional regularization strategies. These improvements significantly strengthen the protection mechanism, enhancing its robustness and effectiveness in safeguarding against unauthorized use and modification of model-generated content.
\cite{peng2025intellectual}

\subsection{Effectiveness of Watermarking}

In general, model watermarking involves a trade-off between image quality and watermark effectiveness.
Many techniques have been proposed to improve the detectability of watermarks while maintaining smaller image perturbations.
WOUAF \cite{kim2024wouaf} embeds watermark messages into generated images by modulating the parameters of each layer within the LDMs' decoder to enhance the effectiveness of watermarks. 
RingID \cite{ci2024ringid} proposes a multi-channel heterogeneous watermarking approach to promote the accuracy in multi-key indentification.
At the same time, in the context of image-based active defense watermarking such as advDM \cite{liang2023adversarial} or Glaze \cite{shan2023glaze}, enhancing the effectiveness of proactive watermarks is also crucial.
Since the diffusion model itself is a denoising process, proactive watermarking methods that rely on high-frequency components are often ineffective.
Anti-DreamBooth (Anti-DB) \cite{van2023anti} leverages noise optimization on the surrogate diffsuion models to achieve a better effectiveness against DreamBooth.
SIMAC \cite{wang2024simac} and MetaCloak \cite{liu2024metacloak} improved Anti-Dreambooth by incorporating meta-learning techniques and introducing additional regularization methods, which increase the robustness and effectiveness of the protection mechanism, ensuring stronger safeguards against unauthorized use and modification of model-generated outputs.

\subsection{Robustness of Watermarking}
Most watermarking methods achieve watermarking by adding noise to images. However, the magnitude of these noises is usually small. Although these methods exhibit robustness against certain image space transformations, they are susceptible to removal by denoising methods such as DiffPure \cite{nie2022diffusion} or GrIDPure \cite{zhao2024can} and augmentation techniques, leading to watermark failure.
To enhance the robustness of these watermarking methods, numerous advanced techniques have been developed.
Tree-Ring \cite{wen2023tree} introduces Fourier transform and inverse transform to embed the watermark in the frequency domain, thereby preventing spatial-domain denoising and augmentation from affecting the effectiveness of the watermark.
ROBIN \cite{huangrobin} leverages the optimized adversarial noises with the optimal prompt signal to augment the robustness of the watermarking method.
Shallow Diffusion \cite{li2024shallow} leverages a low-dimensional subspace to ensure that the model watermark resides in the null space of this subspace, thereby enhancing the robustness of the watermark.

\section{Challenges and Open Issues} \label{challenge}

Diffusion watermarking is a crucial research area in diffusion model security and privacy. It helps protect data and the intellectual property of models while also serving as a privacy-preserving technique to prevent private data from being edited or manipulated by diffusion models. However, noise-based watermarking still faces several challenges, including high computational costs, robustness against image denoising and augmentation processes, and the development of multi-user and provable model watermarking methods.

\subsection{Computational Efficiency of Model Watermarking}
Model watermarking is typically computationally expensive because it requires adding an image-specific watermark to each image. However, for model passive watermarking, which is used to detect the intellectual property of either images or models, this issue is effectively addressed by embedding the watermark directly into the diffusion training process. As a result, there is no additional computational overhead, making it a more efficient solution.
However, in the context of model-based active defense, the watermark for each sample is generated through multiple rounds of PGD optimization, which consumes significant time and computational resources. Although methods like the small-timestep watermark generation in SimAC \cite{wang2024simac} can improve efficiency, this issue still persists.
A feasible solution is to introduce an adversarial generator and train the watermark generator using adversarial loss. 
This approach can improve the efficiency of watermark generation and significantly reduce the computational cost of proactive watermarking.

\subsection{Attack against Watermarks}

First and foremost, it is important to clarify that we consider watermarking as a protective measure by the model owner to safeguard both data content and the model itself. Therefore, as a form of defense, any attempt to remove these watermarking effects is regarded as an attack.
Model watermarking is similar to adversarial noise in that both involve adding small perturbations to images. In watermarking, these perturbations introduce detectable traces into the image, allowing a pre-trained detector to identify whether the data or model originates from a specific diffusion model based on these watermark-induced perturbations.
Since diffusion models inherently possess strong denoising and generative capabilities, they can be trained to remove these small, artificially added noise patterns from images.
DiffPure \cite{nie2022diffusion}, as a highly successful test-time adversarial defense method, reveals the vulnerability of existing adversarial noise. It effectively removes the impact of adversarial noise while preserving image quality.
Similarly, GrIDPure \cite{zhao2024can} reviews adversarial watermarking challenges, benchmarking fine-tuning approaches and highlighting the vulnerability of most model watermarking methods to denoising models.
To address the vulnerability of watermarks, frequency-domain methods such as Tree-Ring provide valuable insights. Additionally, applying $L_{0}$ or $L_{1}$ norm noise regularization to introduce significant perturbations to individual image pixels can effectively mitigate the impact of denoising methods. Beyond pixel space, model owners can also enhance watermark effectiveness by modifying image content at the semantic level through latent space shifts. However, overall, the robustness of existing watermarking methods still requires further research.

\subsection{Multi-Attribution Watermarks}
Integrating model watermarking with diffusion models can help detect generated content and verify whether it originates from a specific diffusion model. At the same time, precisely identifying individual users is another important application of model watermarking, enabling specific user attribution.
Methods such as HiDDeN \cite{zhu2018hidden} or Stable Signature \cite{fernandez2023stable} can only determine whether data originates from a specific model but cannot identify individual users because they use a binary classifier, which can only verify whether the data comes from a specific model but cannot distinguish between different users.
Methods such as Gaussian Shading \cite{yang2024gaussian} and ConceptWM \cite{lei2024conceptwm} employ multi-bit encoding for watermarking. This allows each user to have a unique code, enabling the identification of the specific origin of an image. Additionally, multi-bit watermarking supports provable watermarking, thereby enhancing the confidence level of watermark detection.
RingID \cite{ci2024ringid} also proposes a multi-channel heterogeneous watermarking approach to promote accuracy in multi-key indentification.


\section{Applications of Visual Watermarking with Diffusion Models} \label{application}

As generative workflows become more complex, traditional watermarking faces challenges in maintaining robustness and traceability. Here, we briefly discuss how these diffusion watermarking make a difference for copyright protection and forensic analysis in practical scenarios. We first introduce how the previously mentioned passive watermarking methods are applied for intellectual property protection and forensic traceability. Then followed by examples of proactive watermarking applications aimed at enhancing data privacy and security, preventing unauthorized use of customized outputs.

\subsection{Copyright Protection and Forensics}
Image publishers can embed watermarks into their released visual contents by steganographic to ensure subsequent forensic information extraction. 
This has traditionally been the primary application scenario for watermarking. However, with the rise of large-scale text-to-image (T2I) models, watermarked images now face threats that extend beyond simple image preprocessing or more sophisticated end-to-end network-based tampering. These images may even undergo an entire training-inference pipeline, where they are used to generate entirely new content, posing unprecedented challenges to watermark robustness and traceability.

Theoretically, given the data-driven nature of generative models, watermarking methods designed for data protection can also be applied to safeguard model intellectual property (IP). By embedding model-specific watermarks uniformly into the training data, the protection watermark may naturally become prior knowledge encoded within DMs' parameters, and Recipe introduces \cite{zhao2023recipe} such a scheme.
However, in practical scenarios, this straightforward generalization necessitates careful consideration of the computational overhead associated with embedding watermarks into large-scale training datasets, as well as the potential degradation in image quality. Interference from previously watermarked training data (by image publishers) may also introduce further challenges.

A natural idea is to ensure that all of the DMs' outputs contain embedded watermarks that can be utilized for forensics. 
WOUAF \cite{kim2024wouaf} embeds watermark messages into generated images by modulating the parameters of each layer within the LDMs' decoder. ENDE \cite{xiong2023flexible} similarly incorporates the messages into the decoder's intermediate outputs.
Stable Signature \cite{fernandez2023stable} employs classic HiDDeN \cite{zhu2018hidden} and fine-tunes LDMs' decoder to embed a binary signature. While the aforementioned methods primarily focus on LDMs' decoder, which is the closest part to the output, Tree-Ring \cite{wen2023tree} takes a deeper step into the denoising phase by introducing ring patterns in the Fourier space of the initial noise vector. This enables the embedding of watermark information at the distributional level rather than merely at the sample level. Later, RingID \cite{ci2024ringid} identifies Tree-Ring's limitations in multi-key recognition capabilities and integrates diverse watermark strengths by multi-channel heterogeneous watermarking.

On the other hand, the unique appeal of model IP protection lies in its ability to facilitate forensic analysis with solely partial outputs watermarked, where watermarked outputs are generated only in response to specific trigger prompts, following the principles of backdoor attacks \cite{chen2017targeted}.
In addition to watermarking unconditional and class-conditional DMs, Recipe \cite{zhao2023recipe} also proposes fine-tuning pre-trained DMs using trigger pairs; however, the chosen trigger prompts lack flexibility. Another approach involves NaiveWM and FixedWM \cite{liu2023watermarking}. NaiveWM embeds a designated keyword at a random position within the prompt, introducing variability in watermark placement. In contrast, FixedWM inserts the specified keyword at a fixed, predetermined location within the prompt, ensuring consistent watermark positioning.

\subsection{Proactive Privacy Protection}
Essentially, watermarking can be understood as a form of steganography, which involves embedding specific information within digital content without significantly affecting the content itself. In traceability and copyright protection, such embedded information is \textbf{input-agnostic}, meaning that the concealed message intended by the watermarking entity is independent of the information in the image at the pixel level. 
More broadly, when this concept is extended to be \textbf{input-specific}, the protection no longer requires manually designing the steganography message. Consequently, the watermark decoder is no longer necessary to recover such messages. Instead, the protective mechanism is directly evaluated through human visual perception.

A widely accepted concept is that visually distorted outputs indicate the presence of protective interventions, fundamentally aiming to counteract DMs, which can be achieved through adversarial attacks or GANs.
In this context, like adversarial watermarking, can serve as a potential solution for proactive privacy protection. 
This concept of disrupting generative models can be traced back to the advent of GANs when prior studies \cite{wang2020deceiving,huang2022cmua} investigated methods to manipulate the outputs of GAN-based image translation and modification models.

The earliest attempt to attack SD, known as AdvDM \cite{liang2023adversarial}, utilizes Projected Gradient Descent (PGD) \cite{madry2017towards} to maximize the training loss of the Latent Diffusion Model (LDM) \cite{rombach2022high} via Monte Carlo estimation. This approach is later refined in Mist \cite{liang2023mist}, which integrates textural loss to enhance attack efficacy. 
Meanwhile, Glaze \cite{shan2023glaze} aims to protect artistic intellectual property by perturbing the encoding process of SD, whereas PhotoGuard \cite{salman2023raising} mitigates unauthorized image inpainting by simultaneously targeting both the encoder and UNet. 
UDP \cite{zhao2023unlearnable} emphasizes the diffusion coefficient's effects across different timesteps and accordingly optimizes the timestep sampling strategy.
SDS \cite{xue2023toward} incorporates score distillation sampling 
into PGD's objective function to enhance attack precision.
For a data-free setting, DUAW \cite{ye2024duaw} employs Visual Language Models (VLMs) to enable broader protection mechanisms. In contrast, Zhu et al. \cite{zhu2024watermark} utilize a GAN-based generator instead of conventional PGD to generate adversarial samples embedded with traceable watermarks. 
Anti-DreamBooth (Anti-DB) \cite{van2023anti} specifically targets DreamBooth \cite{ruiz2023dreambooth} fine-tuning by introducing an innovative backpropagation surrogate that learns from both clean and partially adversarial examples. Expanding upon Anti-DB, SimAC \cite{wang2024simac} introduces an adaptive greedy time interval selection strategy to enhance the precision of distortions. Lastly, GrIDPure \cite{zhao2024can} provides a comprehensive review of adversarial watermarking challenges, offering a benchmark analysis of various fine-tuning approaches.

\section{Conclusion and Future Directions} \label{conclusion}

Diffusion models demonstrate immense potential in the field of visual watermarking, providing superior robustness and adaptability. These models are capable of producing high-quality watermarks that are highly resistant to attacks or distortions, thereby safeguarding the integrity of digital content. In this survey, we review both passive and proactive diffusion watermarking techniques, as well as their effectiveness and robustness. Additionally, current diffusion watermarking methods face challenges such as computational efficiency, vulnerability to attacks, and the need for multi-attribute localization and interpretability.
As the demand for secure digital content increases, the innovative capabilities of diffusion models are poised to play a crucial role in advancing watermarking technologies, ultimately enhancing copyright management and privacy protection.

\subsection{Emerging Trends}

The future of watermarking technology integrated with diffusion models will likely advance in various key directions. First, combining diffusion models with vision-language models (VLMs) could enhance watermark robustness against digital manipulation while preserving content fidelity. Then, improving watermark interpretability is crucial. This includes tracing embedded patterns and identifying ownership origins. These advancements will help solve attribution challenges and strengthen accountability in digital content management. Otherwise, optimizing algorithms for imperceptibility and computational efficiency will ensure adaptability to evolving digital threats. The emergence of lightweight techniques is likely to facilitate more accessibility and efficiency for a broader range of use cases.

Addressing AI-generated content authentication and establishing standardized protocols will be critical for maintaining trust in digital ecosystems. Cross-domain collaboration among researchers, industries, and policymakers will drive innovations in watermarking frameworks. Looking ahead, future trends in watermarking technology may also integrate distributed learning approaches, such as federated learning, to enhance watermark generation and protection. By leveraging decentralized data training, these methods can improve watermark robustness while preserving data privacy and security. This approach could further mitigate the risks of adversarial attacks and unauthorized tampering. As generative AI continues to evolve, the combination of distributed learning, explainable AI, and standardized protocols will pave the way for more adaptive, secure, and transparent watermarking solutions, ultimately supporting the integrity of digital content in an increasingly interconnected world.

\subsection{Implications and Challenges}
Diffusion watermarking offers robust solutions to safeguard content in the intelligent digital era. It ensures the authenticity and traceability of digital content and acts as a powerful privacy-preserving mechanism, preventing unauthorized editing or manipulation of sensitive data by diffusion models. Furthermore, the integration of diffusion models with watermarking technologies enhances the resilience of watermarks against adversarial attacks, making them more robust in the face of evolving digital threats. As the digital landscape continues to evolve, the combination of diffusion models and visual watermarking will play a critical role in addressing the dual challenges of content security and privacy preservation.

Besides the challenges mentioned in Section \ref{challenge}, the implementation of real-time watermarking and embedding faces significant technical challenges, particularly in large-scale processing scenarios in the era of generative AI. As AI-generated content becomes more pervasive, the necessity for robust regulatory measures and digital copyright protection grows increasingly urgent. The complexity of monitoring and enforcing watermarking standards in the dynamic environment highlights the challenges of scalability and governance. These factors emphasize the need for adaptable, resilient watermarking solutions that not only address technical limitations but also align with evolving regulatory and technological landscapes, ensuring the integrity and security of digital content in an era dominated by generative AI.

In summary, we anticipate that diffusion watermarking methods will play a crucial role in enhancing digital content security and integrity. As we navigate an increasingly complex media landscape, the ability to effectively embed and detect watermarks will be essential. By understanding the nuances of how these methods function and their potential applications, we can better safeguard intellectual property and foster trust in digital communications.

\section*{Acknowledgments}

This work is partially funded by the National Natural Science Foundation of China (Grant Nos. U21B2045, U20A20223,
32341009, 62206277).

\newpage

\bibliographystyle{named}
\bibliography{ijcai24}

\end{document}